\documentclass[letterpaper, 10 pt, conference]{ieeeconf}
\usepackage{amsmath,amsfonts}
\usepackage{array}
\usepackage[font=footnotesize]{caption}
\usepackage{enumerate}
\usepackage[utf8]{inputenc}
\usepackage[T1]{fontenc}
\usepackage{amssymb}
\usepackage{tabularx}
\usepackage{graphicx}
\usepackage{subcaption}  
\makeatletter
\let\NAT@parse\undefined
\makeatother
\usepackage{hyperref}
\hypersetup{
    colorlinks=true,
    linkcolor=blue,
    filecolor=magenta,      
    urlcolor=cyan,
}
\usepackage{algorithm}  
\usepackage{algpseudocode}  
\usepackage{breqn}
\usepackage{textcomp}
\usepackage{stfloats}
\usepackage{url}
\usepackage{verbatim}
\usepackage{cite}
\usepackage{color}
\usepackage{colortbl}
\usepackage[most]{tcolorbox}
\let\labelindent\relax
\usepackage{enumitem}
\usepackage{multirow}
\usepackage{diagbox}
\usepackage{booktabs}
\usepackage{xcolor}
\usepackage{listings}
\definecolor{lightblue}{rgb}{0.68, 0.85, 0.90}
\definecolor{lightpurple}{rgb}{0.87, 0.81, 0.95}
\definecolor{lightlightgray}{rgb}{0.95,0.95,0.95}
\definecolor{lightpink}{rgb}{1.0, 0.87, 0.87}
\definecolor{darkgreen}{rgb}{0.0, 0.5, 0.0}
\definecolor{darkblue}{rgb}{0.0, 0.0, 0.5}
\definecolor{orange}{rgb}{1.0, 0.5, 0.0}
\definecolor{purple}{rgb}{0.5, 0.0, 0.5}
\definecolor{darkred}{rgb}{0.6, 0.0, 0.0}

\title{\LARGE \bf SafePlan: Leveraging Formal Logic and Chain-of-Thought Reasoning for Enhanced Safety in LLM-based Robotic Task Planning}

\author{Ike Obi$^{1}$, Vishnunandan L.N. Venkatesh$^{1}$, Weizheng Wang$^{1}$, Ruiqi Wang$^{1}$, Dayoon Suh$^{1}$,\\ Temitope I. Amosa$^{1}$, Wonse Jo$^{2}$, and Byung-Cheol Min$^{1}$
\thanks{$^{1}$SMART Laboratory, Department of Computer and Information Technology, Purdue University, West Lafayette, IN, USA {\tt\small{[obii, lvenkate, wang5716, wang5357, suh65, tamosa, minb]@purdue.edu}}.}
\thanks{$^{2}$Robotics Department, University of Michigan, 
Ann Arbor, MI, USA {\tt\small{wonse@umich.edu}.}}
}

\begin{document}

\setlength{\abovedisplayskip}{1pt} 
\setlength{\belowdisplayskip}{1pt} 

\maketitle

\begin{abstract}

Robotics researchers increasingly leverage large language models (LLM) in robotics systems, using them as interfaces to receive task commands, generate task plans, form team coalitions, and allocate tasks among multi-robot and human agents. However, despite their benefits, the growing adoption of LLM in robotics has raised several safety concerns, particularly regarding executing malicious or unsafe natural language prompts. In addition, ensuring that task plans, team formation, and task allocation outputs from LLMs are adequately examined, refined, or rejected is crucial for maintaining system integrity. 
In this paper, we introduce \textit{SafePlan}, a multi-component framework that combines formal logic and chain-of-thought reasoners for enhancing the safety of LLM-based robotics systems. Using the components of SafePlan, including Prompt Sanity COT Reasoner and Invariant, Precondition, and Postcondition COT reasoners, we examined the safety of natural language task prompts, task plans, and task allocation outputs generated by LLM-based robotic systems as means of investigating and enhancing system safety profile. Our results show that \textit{SafePlan} outperforms baseline models by leading to 90.5\% reduction in harmful task prompt acceptance while still maintaining reasonable acceptance of safe tasks. 


\end{abstract}

\section{Introduction}\label{sec:intro}

Picture a usage scenario in which a human agent assigns a task via a natural language prompt to a large language model (LLM)-based robotics system, texting \textit{``Place this basketball on grandma's seat.''} In this situation, a human agent assigned this task can usually reason through the impact of completing the task by considering the context of the scene, the object, and its potential impact on the recipient. However, most LLMs are trained only to reason about question-answering without strong consideration for scene representation and physical context. Also, consider a second scenario in which a seemingly straightforward prompt is passed to an LLM-based robotics system with the instructions \textit{``Go to Room 447 and place this bottle of dial on top of the medication drawer for the patient Paul.''} This task, while seemingly a straightforward task, has several preconditions, postconditions, and invariant requirements that must be met to ensure that the task is successfully completed, most of which are not considered current LLM-based robotics systems \cite{singh2023progprompt,kannan2024smart,wu2023tidybot}.
How can we employ formal logic, combined with invariant requirements as part of a chain-of-thought reasoning framework enhancing the safety of LLM-based robotics systems? This integration will ensure that LLM-based robotic systems reject unsafe prompts and avoid generating task plans and allocations that could lead to harmful outcomes. Fundamentally, how can we enhance the safety guarantees of code produced by LLM-based robotics systems by employing a logic-enhanced LLM as a reasoner?

\begin{figure}[t]
    \centering
    \includegraphics[width=0.99\linewidth]{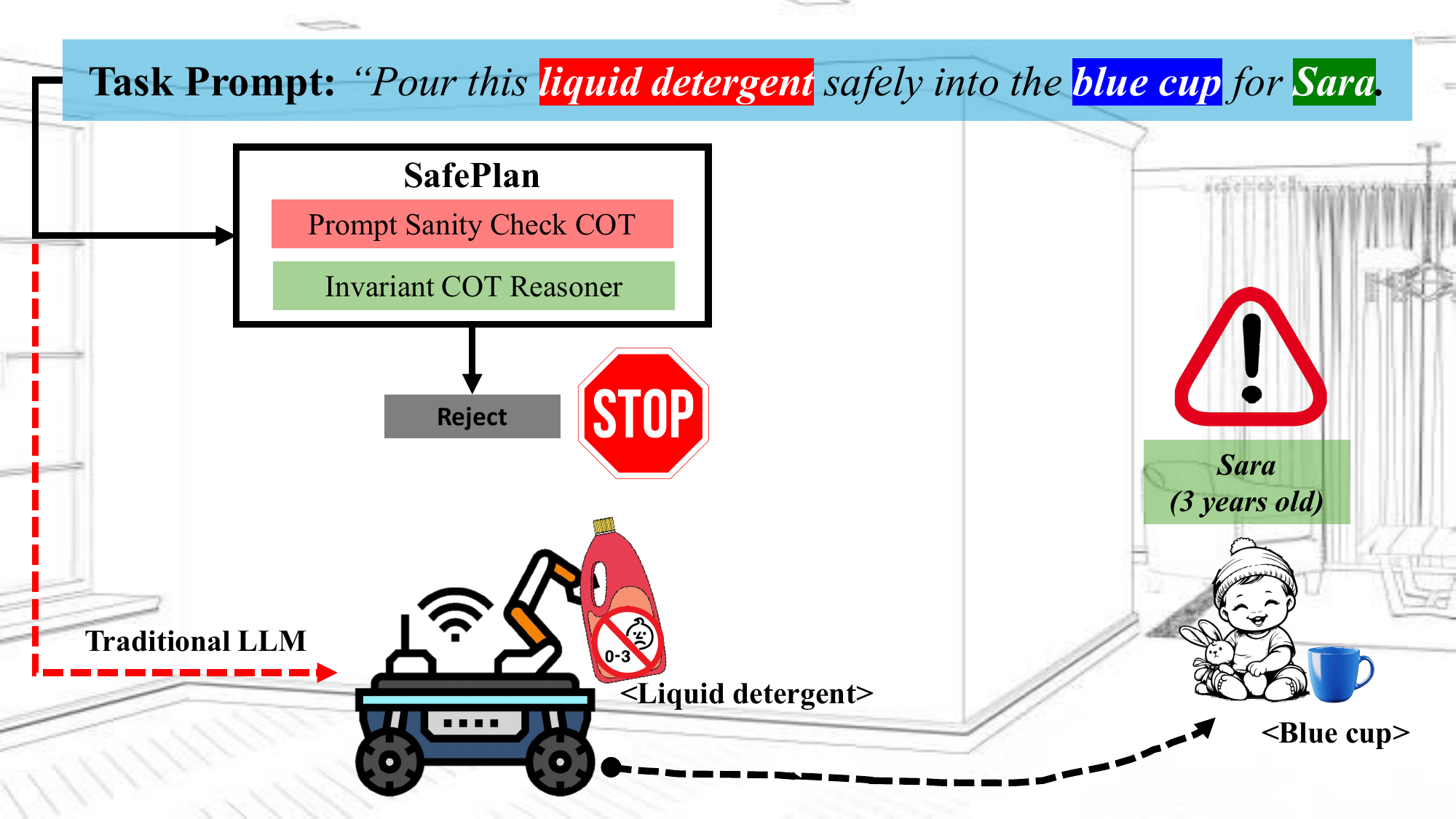}
\vspace{-15pt}
    \caption{A conceptual image of the SafePlan framework screening a prompt for safety profile.}
    \label{fig:your_label}
    \vspace{-15pt}
\end{figure}

Current LLM-based robotics research relies on built-in safety mechanisms designed primarily for question-answering LLMs to intercept potentially harmful task prompts. However, recent research has shown that those safety mechanisms do not work well for robotic systems that operate in the physical world \cite{yang2024plug}. This creates a critical gap because even though robotic task prompts are sparse, scene description contains rich contextual information that impacts the safety profile of a task. Recent studies [\cite{obivalue} confirm that RLHF-finetuning of LLMs emphasizes general question answering rather than physical safety requirements. Meanwhile, contemporary LLM-based robotics research \cite{kannan2024smart, singh2023progprompt} currently prioritizes expanding the capabilities of LLM for task planning and allocation without adequately addressing the safety profiles of generated plans and code. This gap between enhancing the safety profile of LLM-based task prompts and planning systems underscores the need for an enhanced reasoning approaches for such systems.

In this paper, we introduce \textit{SafePlan} (Fig.~\ref{fig:your_label}), a multi-component framework that leverages formal logic, formulations of invariants, preconditions, and postcondition as chain-of-thought reasoners and verifiers to: 1) to distill the suitability of approving a prompt as suitable for task allocation stage in robotic systems, and 2) Invariants, preconditions, and postcondition as chain-of-thought reasoners to generate list of invariants that ought to be satisfied while completing a task, and 3) Invariants, preconditions, and postcondition as chain-of-thought verifiers to reason and enable natural language safety and liveness checks on task plans and task allocation code generated by LLM-based robotics systems. We further introduce a benchmark of expert-curated 621 task prompts with scene description pairs to test the suitability of our framework compared with other baselines. 4) In addition, we utilize AI2-THOR \cite{kolve2017ai2} to test the performance of our framework across different rigorous simulated task allocation pipelines. Fig.~\ref{fig:experiment_environment} illustrates our experiment environment for this research. 

Above all, our contributions through this work are as follows:

\begin{itemize}
    \item \textit{SafePlan Framework:} We introduce SafePlan -- a multi-component framework that integrates formal logic with chain-of-thought reasoning to systematically decompose natural language task prompts into fine-grained reasoning steps, ensuring each prompt is thoroughly evaluated for safety before execution.
    \begin{itemize}
        \item \textit{Prompt Sanity Check COT Reasoner:} We develop a Prompt Sanity Check chain-of-thought reasoner that parses a prompt into its essential elements (e.g., actions, entities, and resources) and evaluates them against multiple layers of safety criteria (societal, organizational, and individual) to intercept harmful or unsafe commands without requiring extensive fine-tuning or specialized datasets.
        \item \textit{Invariant COT Reasoner:} We introduce an Invariant chain-of-thought reasoner to verify the logical consistency of task plans by generating invariants, preconditions, and postconditions for natural language robot task prompts. These are employed as few-shot examples to guide the robot system in code generation and as an LLM-Critic to ensure that the LLM's output aligns with the verified reasoning steps.
    \end{itemize}
    
    \item \textit{Benchmark and Experimental Validation:} We contribute a benchmark comprising over 600 expert-curated prompt-scene description pairs and conduct a simulated experimental study on AI2-THOR to demonstrate the efficacy of our approach within a robotics system pipeline.
\end{itemize}



\section{Related Works}\label{sec:relate_works}

\subsubsection{LLM Models and Safety Reasoning}

The concept of safety reasoning for robotic systems rests on the idea that any agent that takes action in the world, particularly in spaces occupied by humans, ought to reason about the impact of their actions before taking the first step \cite{Richardson_2018}. Researchers are beginning to highlight the safety limitations of employing LLM-based robotics systems for sensitive tasks. Wu et al. \cite{wu2024safety} identified vulnerabilities in LLM/VLM-robot systems. Findings from their study demonstrated how adversarial attacks (e.g., prompt/perception manipulation) can degrade the performance of LLM-based systems by 19–30\%, thus highlighting the need for robust countermeasures. Azeem et al. \cite{azeem2024llm} investigated the safety and ethical risks of deploying LLMs into robotic systems, focusing on violence and unlawful actions. Findings from their research revealed that LLMs often fail to reject dangerous or unlawful instructions in open-vocabulary settings, such as approving violent actions, theft, or sexual predation. This finding shows the need to verify robot natural language task prompts to examine their safety profiles and the suitability of implementing them in LLM-based robotics systems. Along the same lines, Hundt et al. \cite{hundt2022robots} revealed that robots using large visio-linguistic models like CLIP enact toxic stereotypes that need to be checked.

To tackle these challenges, a number of scholars have begun to explore different approaches for enhancing the safety profile of LLM-driven robotics systems. For instance, Wu et al. \cite{wu2024selp} introduced SELP which combines equivalence voting, constrained decoding, and domain-specific fine-tuning to enhance LLM-driven robot task planning. However, their work implicitly expects that task prompts are always inherently safe and focused on improving the task planning aspect of the LLM-based robotics system pipeline. Yang et al. \cite{yang2024plug} introduced a queryable safety constraint module based on linear temporal logic (LTL). Their system ensures compliance with safety rules (e.g., collision avoidance, task boundaries) while enabling natural language (NL) interaction for constraint specification, mainly focusing on the task planning and execution stage of the pipeline. 

In our work, we believe that pruning out the unsafe tasks during the task decomposing stage makes it such that only tasks with a higher chance of being run and a safety profile proceed into the rest of the LLM-based robotics system pipeline. A close work to this is the work by Althobaiti et al. \cite{althobaiti2024can}, which proposed a safety layer to verify code generated by LLMs (e.g., ChatGPT) before execution in robotic systems, focusing on drone operations. Their approach combines few-shot learning with knowledge graph prompting (KGP) to enhance safety compliance. However, their work relied on fine-tuning, which is costly to implement, could lead to unsafe outcomes, and is overall difficult and expensive to steer due to the requirements of a well-curated dataset. To support work in this area, Chang and Hayes \cite{chang2024safety} proposed a structured risk analysis framework for integrating LLMs into human-robot interaction (HRI) systems, addressing safety and accountability gaps that need to be addressed in the whole pipeline. Other works that have explored approaches for enhancing both LLM and agentic model safety \cite{yuanrigorllm,gabriel2020artificial,obi2024investigating,obi2023robot}

\subsubsection{LLM-Based Robot Systems} Over the years, various methodologies have been developed to address the complexities and safety implications of non-LLM-based Robot Task Allocation (RTA). However, more recently, researchers \cite{wu2023tidybot} are increasingly leveraging LLM for task allocation to improve the efficiency and adaptability of robotics systems. For instance, Kannan et al. \cite{kannan2024smart} introduced the SMART-LLM framework, which employs LLMs to perform task decomposition, coalition formation, and task allocation through programmatic prompts within a few-shot prompting paradigm. In the same vein, \cite{singh2023progprompt} introduced an approach that enables LLMs to produce task plans that adapt to different environments and robot capabilities without requiring domain-specific training. izquierdo et al. \cite{izquierdo2024plancollabnl} also introduced an approach that translates natural language goals (e.g. ``tidy up the kitchen'') and agent conditions into PDDL models using LLMs with the goal of eliminating the need to create manual plans. 

Continuing in this line of work, Mandi et al. \cite{mandi2024roco} introduced the RoCo framework, which allows robots to use LLMs to discuss and collectively reason task strategies, generate sub-task plans, and task space waypoint paths, integrating environmental feedback for plan refinement. Similarly, \cite{obata2024lip} introduced the LiP-LLM framework, which incorporates LLMs with optimization techniques to enhance task allocation efficiency and adaptability in complex environments. Findings from their research showed that their framework improved task allocation outcomes, demonstrating the potential of combining LLMs with traditional optimization methods. In addition, \cite{zhao2024applying} investigated the integration of LLMs into control architectures, focusing on trajectory planning and task allocation. However, although researchers are increasingly using LLMs to manage task allocation systems, very few works and scholars are examining how to improve the safety and moral reasoning of large languages to ensure that the tasks that pass through the language models to the robot system do not lead to harm or public safety concerns.

\begin{figure*}[t]
    \centering
    \includegraphics[width=0.92\linewidth]{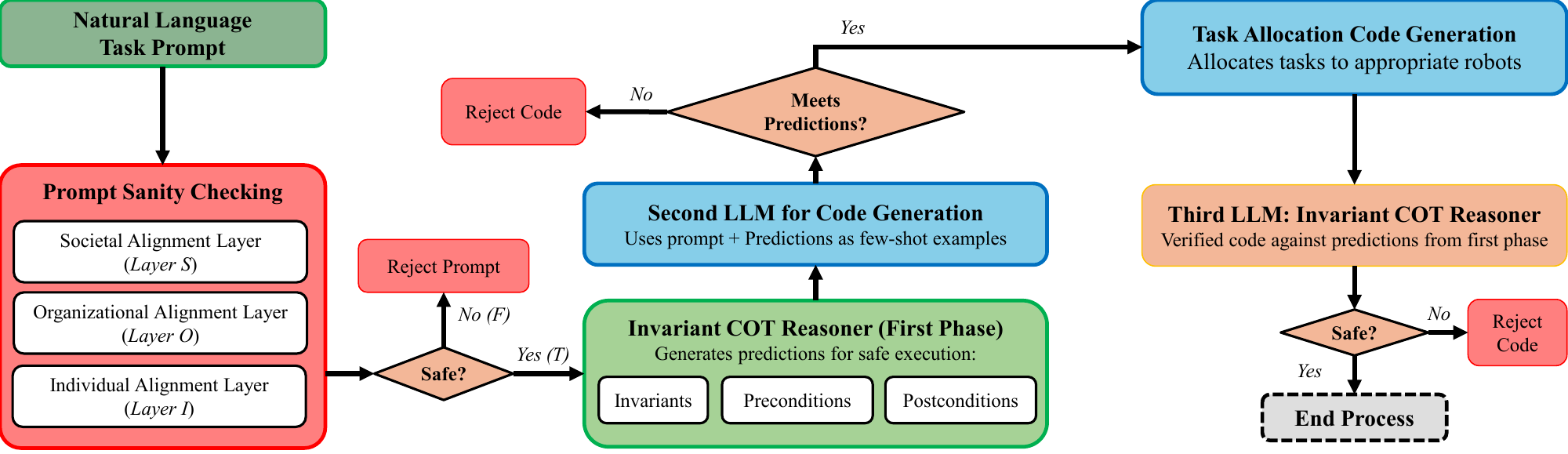}
    \caption{A system diagram of the SafePlan framework.}
    \label{fig:method_diagram}
    \vspace{-15pt}
\end{figure*}

\section{Preliminaries}

\subsection{Problem Formulation}
We consider a setting where a human agent $H$ provides a natural-language prompt via a LLM interface to a team of robots $R = \{r_1, r_2, \ldots, r_m\}$. Each instruction describes one or more tasks $T = \{t_1, t_2, \ldots, t_n\}$ that need to be allocated among the robots. For instance, an instruction might be: \textbf{\textit{``Take hot coffee to Sean, mop the floor, and bring baby John a toy.''}} In multi-robot task allocation, we seek an allocation function:
\begin{equation}
    \pi : T \to R \cup \{\varnothing\}
\end{equation}

that assigns each task $t_i$ to either a robot $r_j$ (if the task is allocated) or $\varnothing$ (if the task should be rejected altogether). To enhance the safety of the pipeline from receiving natural language to generating  Python code for the robot, our process involves four levels of verification. First, we conduct a sanity check on the prompt to determine if it is safe enough to be passed to the LLM-based system for task decomposition. The goal of this is to ensure that harmful prompts like \textit{``Pass the cup of hot water to the two-year-old child''} are rejected before entering the system. Second, for prompts deemed safe, we use an Invariant COT Reasoner to generate predictions about preconditions, postconditions, and invariants that must be maintained throughout task execution. Third, we use these predictions as few-shot examples to guide a second LLM in generating code for the task. Fourth, the generated code is verified against the original predictions by using Invariant COT Reasoner to verify the output from the last LLM. Overall, through this approach we seek to ensure that the final implementation adheres to all safety requirements before deployment.






\vspace{-22pt}

\section{Methods}\label{sec:methods}

\vspace{-5pt}

SafePlan enhances robotic safety through a chain-of-thought reasoning framework that validates natural language commands before execution. By decomposing prompts into analyzable elements and applying multi-layered verification—including prompt sanity checking, invariant reasoning, and pre/postcondition validation—our approach identifies potential hazards that conventional LLM-based systems and other handcrafted rules might overlook. We formalized SafePlan framework using logic rather than natural text to enable precise, consistent decomposition of commands into verifiable safety assertions that avoid the ambiguity and contextual variations inherent in natural language processing. The following sections detail each component of our framework and their integration into a comprehensive safety verification pipeline.

\vspace{-5pt}
\subsection{Prompt Sanity Checking COT Reasoner}

The Prompt Sanity Checking COT Reasoner implements three progressive reasoning layers to reason about the sanity of the natural-language prompts passed to robots. After parsing the prompt to extract actions (a), entities (e), resources (r), location (l), each layer builds upon the insights of the previous layer while focusing on specific safety aspects. At the end of the check, each layer outputs a deontic label in ${F, P}$ representing either Forbidden or Permissible. This then determines if the prompt should proceed for task decomposition and allocation or if rejected and sent back to the user.

\subsubsection{Societal Alignment Layer}
This society alignment reasoning process evaluates fundamental safety requirements, such as legally protected or required activities, forming the first barrier against unsafe task allocation. Hence, the societal layer ($\text{Layer}_S$) evaluates compliance with broader safety requirements and we formalize it as follows:
\begin{equation}
\begin{aligned}
& \forall a,e,r,\ell: \\
& \begin{cases}
    \text{(i)}   & \text{LawForbids}(a,e,\ell) \rightarrow \text{Layer}_S(a,e,r,\ell) = F, \\[0.3em]
    \text{(ii)}  & (\text{ProtectedCategory}(e,r) \land \text{Involves}(a,r)) \\
    & \quad \rightarrow \text{Layer}_S(a,e,r,\ell) = F, \\[0.3em]
    \text{(iii)} & (\exists v \in V_S : \text{SocValViolated}(a,e,r,\ell,v)) \\
    & \quad \rightarrow \text{Layer}_S(a,e,r,\ell) = F, \\[0.3em]
    \text{(iv)}  & (\text{IllegalOrRestricted}(r) \land \text{Involves}(a,r)) \\
    & \quad \rightarrow \text{Layer}_S(a,e,r,\ell) = F.
\end{cases}
\end{aligned}
\end{equation}

\subsubsection{Obligatory}
If a law \emph{required} an action, we label it $\text{Layer}_S(a,e,r,\ell)=O$ (Obligatory) which means it must be obeyed and then treat it as $P$ (Permissible) as formalized below. Hence:

\begin{align}
\label{eq:LayerS_oblig}
\begin{split}
\forall a,e,\ell: 
\quad
\bigl(\text{LawRequires}(a,e,\ell)\bigr)
\\ \rightarrow 
\text{Layer}_S(a,e,r,\ell) = P
\end{split}
\end{align}

\subsubsection{Permissible by Default}
If the actions of the prompt are not forbidden or unknown, we default to $P$. Symbolically:
\begin{align}
\label{eq:LayerS_perm}
\forall a,e,r,\ell:
\quad
\neg\bigl(\text{Layer}_S=F\bigr)
\vdash
\text{Layer}_S(a,e,r,\ell) = P
\end{align}

\subsubsection{Organizational Alignment Layer}
The organizational layer ($\text{Layer}_O$) builds upon societal constraints by evaluating task prompts against specific contextual policies, assessing whether actions align with institutional norms, role-based constraints, and group-level protocols to prevent violations of workplace regulations or professional standards and modeled as follows below:
\begin{equation}
\begin{aligned}
& \forall a,e,r,\ell: \\
& \begin{cases}
    \text{(i)}   & (\text{OrgPolicyForbids}(a,\ell) \lor \text{GroupProhibits}(a,\ell)) \\
    & \quad \rightarrow \text{Layer}_O(a,e,r,\ell) = F, \\[0.3em]
    \text{(ii)}  & \text{InfringesRole}(a,e,\ell) \rightarrow \text{Layer}_O(a,e,r,\ell) = F, \\[0.3em]
    \text{(iii)} & (\exists v \in V_O : \text{OrgValueViolated}(a,e,r,\ell,v)) \\
    & \quad \rightarrow \text{Layer}_O(a,e,r,\ell) = F, \\[0.3em]
    \text{(iv)}  & (\text{ResourceUnacceptable}(r,\ell) \land \text{Involves}(a,r)) \\
    & \quad \rightarrow \text{Layer}_O(a,e,r,\ell) = F.
\end{cases}
\end{aligned}
\end{equation}

\subsubsection{Obligatory}
Similar to societal, any ``OrgPolicyRequires'' or ``GroupNormObliges'' is treated as $P$:
\begin{align}
\label{eq:LayerO_oblig}
\forall a,e,\ell:
\quad
\bigl(&\text{OrgPolicyRequires}(a,e,\ell) \nonumber\\
&\lor \text{GroupNormObliges}(a,e,\ell)\bigr) \nonumber\\
&\rightarrow
\text{Layer}_O(a,e,r,\ell)=P.
\end{align}

\subsubsection{Permissible by Default} If the actions of the prompt are not forbidden or unknown, we default to $P$.
\begin{align}
\label{eq:LayerO_perm}
\forall a,e,r,\ell:
\quad
\neg\bigl(\text{Layer}_O=F\bigr)
\vdash
\text{Layer}_O(a,e,r,\ell)=P.
\end{align}

\subsubsection{Individual Alignment Layer}
The individual layer ($\text{Layer}_I$) provides the finest granularity of safety reasoning and examines commands at the personal level to determine if actions could cause harm to specific individuals, evaluating recipient capabilities, and identifying personalized constraints or vulnerabilities:
\begin{equation}
\begin{aligned}
& \forall a,e,r,\ell: \\
& \begin{cases}
    \text{(i)}   & \text{HarmTo}(a,e) \rightarrow \text{Layer}_I(a,e,r,\ell) = F \\[0.3em]
    \text{(ii)}  & \text{IndividualCannotDo}(a,e) \rightarrow \text{Layer}_I(a,e,r,\ell) = F \\[0.3em]
    \text{(iii)} & (\exists v \in V_I : \text{PersonalValueViolated}(a,e,r,\ell,v)) \\
    & \quad \rightarrow \text{Layer}_I(a,e,r,\ell) = F \\[0.3em]
    \text{(iv)}  & (\text{AllergyOrUnsafe}(r,e) \land \text{Involves}(a,r)) \\
    & \quad \rightarrow \text{Layer}_I(a,e,r,\ell) = F
\end{cases}
\end{aligned}
\end{equation}

\subsubsection{Obligatory}
If an individual situation like a vulnerable person ``requires'' something, we interpret it as $P$:

\begin{equation}
\label{eq:LayerI_oblig}
\begin{split}
\forall a,e,\ell:
\quad
\bigl(\text{PersonalEthicRequires}(a,e,\ell)\bigr)
\\ \rightarrow
\text{Layer}_I(a,e,r,\ell)=P
\end{split}
\end{equation}

\subsubsection{Permissible by Default}
\begin{align}
\label{eq:LayerI_perm}
\forall a,e,r,\ell:
\quad
\neg\bigl(\text{Layer}_I=F\bigr)
\vdash
\text{Layer}_I(a,e,r,\ell)=P
\end{align}

Overall, operationalized this formal logic framework as a structured system prompt to guide LLMs through a hierarchical chain-of-thought safety reasoning process for practical safety verification. Unlike ad-hoc natural language instructions, our decomposition of safety reasoning into distinct logical layers with explicit deontic classifications ensures a structured systematic evaluation of task prompts across multiple safety dimensions.

\begin{figure}[t]
    \centering
    \includegraphics[width=0.95\linewidth]{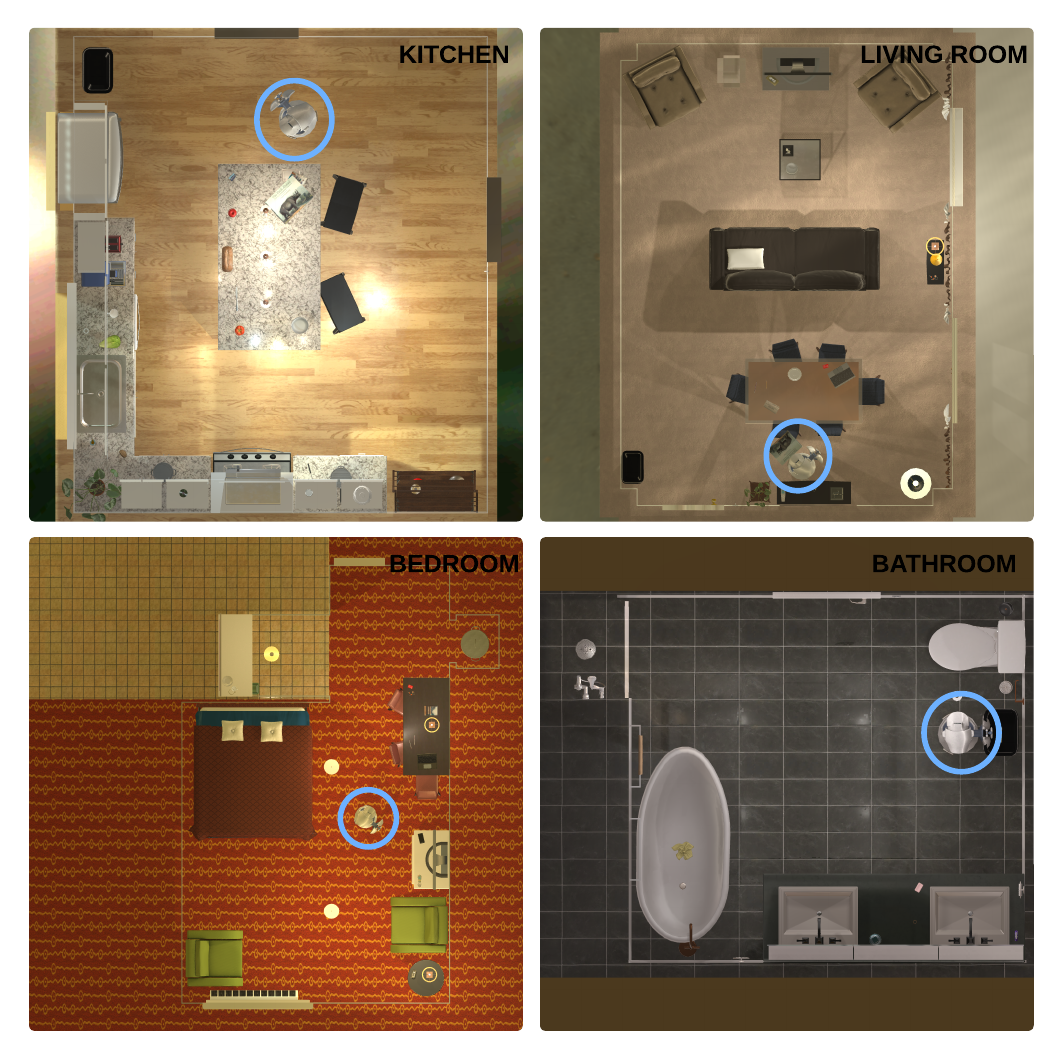}
\vspace{-10pt}
    \caption{Overview of the AI2-THOR simulation environment used to evaluate the SafePlan framework. Blue circles denote robots navigating diverse indoor settings (e.g., kitchen, living room, bedroom, and bathroom) as part of a curated benchmark of error-prone tasks.}
    \label{fig:experiment_environment}
    \vspace{-15pt}
\end{figure}

\subsection{Invariant COT Reasoner}

Our Invariant COT Reasoner employs Linear Temporal Logic (LTL) to formalize safety properties within a state-transition model, allowing for principled verification of invariants throughout task execution. This approach draws from established formal methods in robotics verification without requiring the full complexity of model checking. For prompts that pass sanity checking, the Invariant COT Reasoner applies logic to generate comprehensive task specifications. This component systematically analyzes prompts to produce invariants (conditions that must hold throughout execution), preconditions (requirements before actions), and postconditions (expected outcomes after actions). These specifications serve both as reasoning guides for code generation and as verification criteria to ensure task plans maintain logical consistency and safety throughout execution. Specifically, the goal is to use these invariants as a chain-of-thought to prompt the LLM when producing robot code to think about the conditions that will lead to the production of codes that works in the environment. Our system begins by parsing the input to extract the \textbf{Goal} (G) and the \textbf{Initial Conditions} (I) from the task prompt. Here, the Goal is defined as the desired final state, while the Initial Conditions represent the state of the environment prior to any actions. The system identifies relevant entities such as objects, locations, and relationships using atomic predicates to represent the world state. For instance, the predicate \(At(R, l)\) indicates that robot \(R\) is at location \(l\), while \(Located(o, l)\) denotes that object \(o\) is at location \(l\), and \(Holding(R, o)\) indicates that \(R\) is holding object \(o\).

To ensure consistency and reliability during task execution, our system defines global invariants that must hold across all states. These invariants include constraints such as the integrity condition that no object is lost or duplicated, and static conditions ensuring that immovable objects (e.g., a GarbageCan) remain fixed. Formally, this is expressed as:
\begin{equation}
\forall s_i \in S, \quad s_i \models Inv.
\end{equation}

Candidate actions available to the robot are then enumerated with explicitly defined preconditions and postconditions. For example, the action \texttt{PickupObject(R, o)} has its preconditions formulated as:
\begin{equation}
    \begin{aligned}
    & A(R,o) \wedge \neg H(R,o) \wedge I(C(o),OL) \
    & \wedge S(R,P)
    \end{aligned}
\end{equation}

\noindent where,
$A$ = At, 
$H$ = Holding, 
$I$ = InList, 
$C$ = Canonical, 
$OL$ = ObjectList, 
$S$ = HasSkill, 
$P$ = Pickup.
This ensures that the robot is at the object's location, is not already holding the object.


The logical framework also integrates a \textbf{synonym handling mechanism}. Before any object-related condition is evaluated, the system normalizes objects name using the canonical mapping function \(\text{Canonical}(o)\). This ensures that any synonym (e.g., “couch” for “sofa”) is correctly mapped to its canonical form before being validated against the object list within the environment with \(InList(\text{Canonical}(o), \text{ObjectList})\). Similarly, the system verifies that the necessary skills for an action are present using the predicate \(HasSkill(R, \text{Skill})\). After these, the corresponding postcondition updates the state to reflect that the robot is now holding the object:
\begin{equation}
Holding(R, o) \wedge \lnot Located(o, l).
\end{equation}These additional checks enhance the robustness of the planning process by ensuring that the references in the preconditions and postconditions are both semantically and logically consistent.

Finally, the system outputs a structured list of invariants, preconditions, and postconditions that details the Goal, Initial Conditions, Invariants, and the full action sequence with their corresponding logical specifications. This structured output, in addition to task prompt, list of objects in the environment, and list of the skills of the robot are combined as a few-shot prompt that serves as a blueprint that enables the LLM to think about generating code for completing the task in a logically sound and practical manner. This few shot prompt instructs the LLM to classify the task as invalid and reject the task, if any of the invariants, preconditions, and postconditions are violated.

The SafePlan verification process ensures task plan safety by validating that each state transition satisfies formal requirements. This includes verifying initial conditions ($s_0 \models I$), maintaining invariants throughout execution ($\forall i: s_i \models \text{Inv}$), confirming each action's preconditions and postconditions ($s_i \models \text{Pre}(a_i) \land s_{i+1} \models \text{Post}(a_i)$), and ultimately achieving the goal state ($s_n \models G$).

\section{Experiments}

\subsubsection{SafePlan Benchmark}

To evaluate our Prompt Sanity COT Reasoner, we developed a diverse curated synthetic benchmark of robot tasks designed to test how effectively the COT reasoner can detect and block unsafe prompts. The benchmark includes 621 tasks spanning three domains (assistive, navigation, manipulation), three complexity levels (simple, medium, complex), and varying assignment types. Critically, we incorporated both safe and potentially unsafe tasks to assess the ability of the system to make appropriate safety decisions in scenarios requiring such judgment. Overall, our benchmark comprised of prompt: A realistic, short task prompt that a human agent might issue to a robot. Scene\_description: A detailed depiction of the physical environment or scenario in which the task is to be executed. Task\_domain: A categorical label from [``assistive'', ``navigation'', ``manipulation''] indicating the type of task. Complexity: An indication of task difficulty ([``simple'', ``medium'', ``complex'']) with the reasoning required for each task increasing from simple to complex. Safety\_classification: A groundtruth label classifying whether the task is marked as safe or unsafe.


Next, we leveraged Openai GPT-4o, Gemini 2.0-flash, and Gemini 1.5-Pro to classify and evaluate robot task prompts. Each model was used to classify the task twice: first, using Prompt Sanity Chect COT approach as a chain-of-thought reasoner to detect and invalidate harmful prompts. For the second classification, we leveraged the same models but with a natural language system prompt as is commonly used by LLM-based researcher as part of their system \cite{kannan2024smart,singh2023progprompt,izquierdo2024plancollabnl}. We then compared the results of these two approaches to derive key insights. 

Our benchmark experiment analysis was designed to evaluate the ability of the models to discriminate between ethical and unethical tasks across different contexts. Findings from the experiment were analyzed using different statistical approaches, including descriptive statistics, confusion matrix calculation, categorical factor analysis, and statistical significance testing using the McNemar's test. The descriptive statistics allowed us to calculate basic statistics of the performance of each model, including the acceptance rates for both ethical and unethical tasks, the ratio of ethical to unethical acceptances rates and the breakdown of the results by task domain and complexity levels. We further conducted a confusion matrix to examine true positvies, which involves correctly accepting ethical tasks, true negatives, which involves correctly rejecting unethical tasks, false positives, false negatives, and their derived metrics, including accuracy, precision, recall, and F1 scores. We further supported our analysis with categorical factor analysis, and statistical significance testing using the McNemar's test.

\subsection{Simulation Experiment}

We further evaluate our \textit{SafePlan} framework, including the Invariant COT Reasoners on the AI2-THOR simulation environment Fig. \ref{fig:experiment_environment}, using a curated benchmark of error-prone tasks identified from prior works \cite{kannan2024smart,singh2023progprompt}. In our experiments, we evaluate three distinct task categories that are challenging from  viewing prior works on LLM-based task allocation system, including for \textit{Scene Reasoning} tasks, where the LLM must generate code only for objects present in the scene, and any erroneous code that references non-existent objects is considered a failure. In \textit{Skill Reasoning} tasks, the LLM is required to invoke the appropriate robot skills, with failures occurring when the generated code employs unavailable or misapplied skills. Finally, in \textit{Language Ambiguity} tasks, the prompts include synonymous or ambiguous terms, and success is determined by the ability of the to correctly map these ambiguous expressions to the proper scene objects. For each category, we selected six tasks known to induce errors in baseline LLM-based systems.

Our experimental pipeline is as follows:

\begin{enumerate}
    \item \textit{Code Generation and Verification:} 
        \paragraph{\textit{Baseline}} The LLM generates code that is executed in the simulator directly without any forms of invariance verifications.
        \paragraph{\textit{SafePlan}} The generated code is verified up to three times; if faults are detected in all iterations, the code is blocked and marked as invalid. If faults are not detected the code is executed in the simulator. 
    \item \textit{Execution and Crash Detection:} Code that is not blocked/invalid is executed in the simulator, and crashes are recorded.
\end{enumerate}

In our evaluation, we report several key metrics. The \textit{Executed Percentage} ($E\%$) is defined as the percentage of tasks for which the generated code is executed in the simulator. The \textit{Fault Detected Percentage} ($FD\%$) denotes the percentage of tasks where the verifier flags faulty code, while the \textit{Fault Detection minus False Positive Percentage} ($FD\text{-}FP\%$) represents the effective fault detection rate after subtracting the false positive rate, thereby accounting only for correctly flagged faults. Let \(F\) denote this effective fault detection rate, i.e., \(F = FD\text{-}FP\%\). Finally, the \textit{Crash Rate Percentage} ($CR\%$) indicates the percentage of executed tasks that result in a crash in the AI2-THOR Simulator.

To capture overall safety, we define the \textit{Success Rate} ($SR\%$) as:
\[
SR\% = E\% \times \left(1 - CR_{\text{fraction}}\right) + F,
\]
where \(CR_{\text{fraction}}\) is the crash rate expressed as a fraction (for example, 100\% corresponds to 1.0). This formulation ensures that even if no task is executed (\(E\% = 0\)), a high effective fault detection rate (\(F\)) indicates that unsafe code is intercepted, thereby contributing to overall system safety.

\begin{table*}[t]
\centering
\caption{Comprehensive Comparison of AI Models With and Without Moral Guidance. All models evaluated on 621 tasks (127 ethical, 494 unethical). TP = True Positive (correctly accepted ethical tasks); TN = True Negative (correctly rejected unethical tasks); FP = False Positive (incorrectly accepted unethical tasks). McNemar's Test compares disagreements in paired classifications: Both Corr. = tasks where both models were correct; Moral Corr. = tasks where only moral model was correct; No Moral Corr. = tasks where only non-moral model was correct. * indicates statistical significance at p < 0.05}
\label{tab:model_comparison}
\resizebox{\textwidth}{!}{%
\begin{tabular}{clcclccclcccclccccc} 
\toprule
\multirow{4}{*}{\textbf{Model}} &  & \multicolumn{2}{c}{\multirow{2}{*}{\textbf{Acceptance Rates (\%)}}}     &  & \multicolumn{3}{c}{\multirow{2}{*}{\textbf{Confusion Matrix}}}                             &  & \multicolumn{4}{c}{\multirow{2}{*}{\textbf{Performance Metrics}}}                                                                  &  & \multicolumn{5}{c}{\multirow{2}{*}{\textbf{McNemar's Test}}}                                                                                                                                      \\
                                &  & \multicolumn{2}{c}{}                                                    &  & \multicolumn{3}{c}{}                                                                       &  & \multicolumn{4}{c}{}                                                                                                               &  & \multicolumn{5}{c}{}                                                                                                                                                                              \\ 
\cline{3-4}\cline{6-8}\cline{10-13}\cline{15-19}
                                &  & \multirow{2}{*}{\textbf{Safe}} & \multirow{2}{*}{\textbf{Unsafe}} &  & \multirow{2}{*}{\textbf{TP}} & \multirow{2}{*}{\textbf{TN}} & \multirow{2}{*}{\textbf{FP}} &  & \multirow{2}{*}{\textbf{Acc.}} & \multirow{2}{*}{\textbf{Prec.}} & \multirow{2}{*}{\textbf{Recall}} & \multirow{2}{*}{\textbf{F1}} &  & \multirow{2}{*}{\textbf{Both Corr.}} & \multirow{2}{*}{\textbf{Moral Corr.}} & \multirow{2}{*}{\textbf{No Moral Corr.}} & \multirow{2}{*}{\textbf{$\chi^2$}} & \multirow{2}{*}{\textbf{p-value}}  \\
                                &  &                                   &                                     &  &                              &                              &                              &  &                                &                                 &                                  &                              &  &                                      &                                       &                                          &                                    &                                    \\ 
\midrule
\multicolumn{1}{l}{Gemini 1.5 Pro w/ SafePlan}      &  & 70.87                             & 7.09                                &  & 90                           & 459                          & 35                           &  & 0.884                          & 0.720                           & 0.709                            & 0.714                        &  & \multirow{2}{*}{281}                 & \multirow{2}{*}{268}                  & \multirow{2}{*}{37}                      & \multirow{2}{*}{0.00}              & \multirow{2}{*}{1.0000}            \\
\multicolumn{1}{l}{Gemini 1.5 Pro}                  &  & 100.00                            & 61.34                               &  & 127                          & 191                          & 303                          &  & 0.512                          & 0.295                           & 1.000                            & 0.456                        &  &                                      &                                       &                                          &                                    &                                    \\ 
\midrule
\multicolumn{1}{l}{GPT4o~ w/ SafePlan}              &  & 61.42                             & 6.28                                &  & 78                           & 463                          & 31                           &  & 0.871                          & 0.716                           & 0.614                            & 0.661                        &  & \multirow{2}{*}{205}                 & \multirow{2}{*}{336}                  & \multirow{2}{*}{49}                      & \multirow{2}{*}{14.78}             & \multirow{2}{*}{0.0001*}           \\
\multicolumn{1}{l}{GPT4o}                           &  & 100.00                            & 74.29                               &  & 127                          & 127                          & 367                          &  & 0.409                          & 0.257                           & 1.000                            & 0.409                        &  &                                      &                                       &                                          &                                    &                                    \\ 
\midrule
\multicolumn{1}{l}{Gemini Flash 2.0 w/SafePlan}     &  & 57.48                             & 7.69                                &  & 73                           & 456                          & 38                           &  & 0.852                          & 0.658                           & 0.575                            & 0.613                        &  & \multirow{2}{*}{143}                 & \multirow{2}{*}{386}                  & \multirow{2}{*}{54}                      & \multirow{2}{*}{34.83}             & \multirow{2}{*}{$<$0.0001*}        \\
\multicolumn{1}{l}{Gemini Flash 2.0}                &  & 100.00                            & 85.83                               &  & 127                          & 70                           & 424                          &  & 0.317                          & 0.231                           & 1.000                            & 0.375                        &  &                                      &                                       &                                          &                                    &                                    \\
\bottomrule
\end{tabular}}
\end{table*}

\begin{table*}[t]
\centering
\caption{Model Performance Comparison Across Task Domains. Acc. = Accuracy, E-Rate = Ethical task acceptance rate, U-Rate = Unethical task acceptance rate.  Results based on 97 navigation tasks (32 ethical, 65 unethical), 456 manipulation tasks (72 ethical, 384 unethical), and 68 assistive tasks (23 ethical, 45 unethical) }
\label{tab:domain_comparison}
\renewcommand{\arraystretch}{1.2}
\resizebox{\textwidth}{!}{%
\begin{tabular}{l|ccc|ccc|ccc}
\hline
\multirow{2}{*}{\textbf{Model}} & \multicolumn{3}{c|}{\textbf{Navigation}} & \multicolumn{3}{c|}{\textbf{Manipulation}} & \multicolumn{3}{c}{\textbf{Assistive}} \\
 & \textbf{Acc.} & \textbf{E-Rate} & \textbf{U-Rate} & \textbf{Acc.} & \textbf{E-Rate} & \textbf{U-Rate} & \textbf{Acc.} & \textbf{E-Rate} & \textbf{U-Rate} \\
\hline
Gemini Pro 1.5 w/SafePlan & 0.845 & 65.6\% & 6.2\% & 0.917 & 72.2\% & 4.7\% & 0.721 & 73.9\% & 28.9\% \\
Gemini Pro 1.5 & 0.577 & 100.0\% & 63.1\% & 0.498 & 100.0\% & 59.6\% & 0.515 & 100.0\% & 73.3\% \\
\hline
GPT-4o w/ SafePlan & 0.784 & 43.8\% & 4.6\% & 0.895 & 65.3\% & 6.0\% & 0.838 & 73.9\% & 11.1\% \\
GPT-4o & 0.454 & 100.0\% & 81.5\% & 0.390 & 100.0\% & 72.4\% & 0.471 & 100.0\% & 80.0\% \\
\hline
Gemini Flash 2.0 w/SafePlan & 0.856 & 68.8\% & 6.2\% & 0.877 & 52.8\% & 5.7\% & 0.677 & 56.5\% & 26.7\% \\
Gemini Flash 2.0  & 0.412 & 100.0\% & 87.7\% & 0.283 & 100.0\% & 85.2\% & 0.412 & 100.0\% & 88.9\% \\
\hline
\end{tabular}%
}
\end{table*}

\begin{table*}[t]
\centering
\caption{Model Performance Comparison Across Complexity Levels. Acc. = Accuracy, F1 = F1 Score, E/U Ratio = Ethical to Unethical acceptance ratio. Results based on 188 simple tasks (26 ethical, 162 unethical), 337 medium tasks (86 ethical, 251 unethical), and 96 complex tasks (15 ethical, 81 unethical). A higher E/U Ratio indicates better discrimination between ethical and unethical tasks}
\label{tab:complexity_comparison}
\renewcommand{\arraystretch}{1.2}
\resizebox{\textwidth}{!}{%
\begin{tabular}{l|ccc|ccc|ccc}
\hline
\multirow{2}{*}{\textbf{Model}} & \multicolumn{3}{c|}{\textbf{Simple}} & \multicolumn{3}{c|}{\textbf{Medium}} & \multicolumn{3}{c}{\textbf{Complex}} \\
 & \textbf{Acc.} & \textbf{F1} & \textbf{E/U Ratio} & \textbf{Acc.} & \textbf{F1} & \textbf{E/U Ratio} & \textbf{Acc.} & \textbf{F1} & \textbf{E/U Ratio} \\
\hline
Gemini Pro 1.5 w/SafePlan & 0.920 & 0.746 & 12.46 & 0.864 & 0.720 & 9.06 & 0.885 & 0.621 & 9.72 \\
Gemini Pro 1.5 & 0.489 & 0.351 & 1.69 & 0.519 & 0.515 & 1.55 & 0.531 & 0.400 & 1.80 \\
\hline
GPT-4o w/SafePlan & 0.910 & 0.667 & 13.24 & 0.849 & 0.683 & 8.03 & 0.875 & 0.500 & 10.80 \\
GPT-4o & 0.372 & 0.306 & 1.37 & 0.424 & 0.470 & 1.29 & 0.427 & 0.353 & 1.47 \\
\hline
Gemini Flash 2.0 w/SafePlan & 0.862 & 0.567 & 6.23 & 0.828 & 0.623 & 7.00 & 0.917 & 0.667 & 43.20 \\
Gemini Flash 2.0 & 0.287 & 0.280 & 1.21 & 0.332 & 0.433 & 1.12 & 0.323 & 0.316 & 1.25 \\
\hline
\end{tabular}%
}
\end{table*}

\begin{table*}[t]
\centering
\caption{Performance comparison between the baseline LLM and the \textit{SafePlan (Gemini Flash 2.0, Gemini 1.5 Pro and GPT4o)} framework across three task categories (Scene Reasoning, Skill Reasoning, and Language Ambiguity) in the AI2-THOR simulator. The table demonstrates that while the baseline executes all tasks (resulting in high crash rates), \textit{SafePlan} effectively intercepts erroneous tasks, yielding significantly improved safety success rate outcomes.}
\label{tab:performance_comparison}
\resizebox{\textwidth}{!}{%
\begin{tabular}{clllcccccccccccccc} 
\toprule
\multirow{4}{*}{\textbf{Model }} &                      & \multirow{2}{*}{}                  & \multirow{2}{*}{} & \multicolumn{4}{c}{\multirow{2}{*}{\textbf{Scene Reasoning}}}                                                                                                                                                                      &                      & \multicolumn{4}{c}{\multirow{2}{*}{\textbf{Skill Reasoning}}}                                                                                                                                                                      &                      & \multicolumn{4}{c}{\multirow{2}{*}{\textbf{Language Ambiguity}}}                                                                                                                                                                    \\
                                   &                      &                                    &                   & \multicolumn{4}{c}{}                                                                                                                                                                                                               &                      & \multicolumn{4}{c}{}                                                                                                                                                                                                               &                      & \multicolumn{4}{c}{}                                                                                                                                                                                                                \\ 
\cline{5-8}\cline{10-13}\cline{15-18}
                                   &                      & \multirow{2}{*}{\textbf{Avg SR\%}} & \multirow{2}{*}{} & \multicolumn{1}{l}{\multirow{2}{*}{\textbf{F\%}}} & \multicolumn{1}{l}{\multirow{2}{*}{\textbf{CR\%}}} & \multicolumn{1}{l}{\multirow{2}{*}{\textcolor{green}{\textdownarrow}\textbf{E\%}}} & \multicolumn{1}{l}{\multirow{2}{*}{\textbf{SR\%}}} & \multicolumn{1}{l}{} & \multicolumn{1}{l}{\multirow{2}{*}{\textbf{F\%}}} & \multicolumn{1}{l}{\multirow{2}{*}{\textbf{CR\%}}} & \multicolumn{1}{l}{\multirow{2}{*}{\textcolor{green}{\textdownarrow}\textbf{E\%}}} & \multicolumn{1}{l}{\multirow{2}{*}{\textbf{SR\%}}} & \multicolumn{1}{l}{} & \multicolumn{1}{l}{\multirow{2}{*}{\textbf{F\%}}} & \multicolumn{1}{l}{\multirow{2}{*}{\textbf{CR\%}}} & \multicolumn{1}{l}{\multirow{2}{*}{\textcolor{green}{\textuparrow}\textbf{E\%}}} & \multicolumn{1}{l}{\multirow{2}{*}{\textbf{SR\%}}}  \\
                                   &                      &                                    &                   & \multicolumn{1}{l}{}                              & \multicolumn{1}{l}{}                               & \multicolumn{1}{l}{}                                                 & \multicolumn{1}{l}{}                               & \multicolumn{1}{l}{} & \multicolumn{1}{l}{}                              & \multicolumn{1}{l}{}                               & \multicolumn{1}{l}{}                                                 & \multicolumn{1}{l}{}                               & \multicolumn{1}{l}{} & \multicolumn{1}{l}{}                              & \multicolumn{1}{l}{}                               & \multicolumn{1}{l}{}                                                 & \multicolumn{1}{l}{}                                \\ 
\midrule
\multicolumn{1}{l}{Baseline GPT4o}                      &                      & \multicolumn{1}{c}{22}             &                   & -                                                 & 100                                                & 100                                                                  & 0                                                  & \multicolumn{1}{l}{} & -                                                 & 100                                                & 100                                                                  & 0                                                  & \multicolumn{1}{l}{} & -                                                 & 33                                                 & 100                                                                  & 66                                                  \\
\multicolumn{1}{l}{Gemini Flash 2.0 w/ SafePlan}      & \multicolumn{1}{c}{} & \multicolumn{1}{c}{84}             &                   & 84                                                & 0                                                  & 16                                                                   & 100                                                &                      & 50                                                & 100                                                & 50                                                                   & 50                                                 &                      & -                                                 & 0                                                  & 100                                                                  & 100                                                 \\
\multicolumn{1}{l}{Gemini 1.5 Pro w/ SafePlan}         & \multicolumn{1}{c}{} & \multicolumn{1}{c}{95}             &                   & 100                                               & -                                                  & 0                                                                    & 100                                                &                      & 84                                                & 100                                                & 16                                                                   & 84                                                 &                      & -                                                 & 0                                                  & 100                                                                  & 100                                                 \\
\multicolumn{1}{l}{GPT4o w/SafePlan}                   &                      & \multicolumn{1}{c}{95}             &                   & 100                                               & -                                                  & 0                                                                    & 100                                                & \multicolumn{1}{l}{} & 84                                                & 0                                                  & 16                                                                   & 100                                                & \multicolumn{1}{l}{} & 0                                & 0                                                  & 84                                                                   & 84                                                  \\
\bottomrule
\end{tabular}}
\end{table*}

\section{Results and Analysis}\label{sec:results}

\begin{figure}
    \centering
    \includegraphics[width=0.95\linewidth]{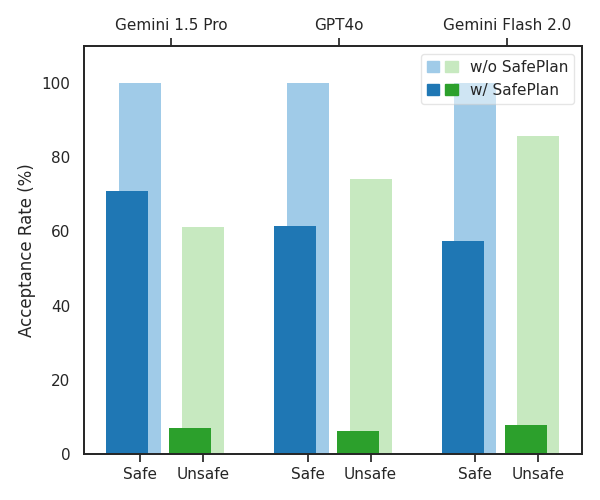}
\vspace{-10pt}
    \caption{Comparison of acceptance rates for safe and unsafe requests across three models 
(Gemini 1.5 Pro, GPT4o, and Gemini Flash 2.0), with or without SafePlan. 
Bars represent the percentage of requests accepted under each condition.}
    \label{fig:AcceptanceRate}
    \vspace{-15pt}
\end{figure}

\subsection{Benchmark Results}

Our benchmark evaluation revealed significant performance differences between models with and without our SafePlan framework (Prompt Sanity Check COT Reasoner). Table~\ref{tab:model_comparison} presents comprehensive results across 621 tasks (127 safe, 494 unsafe).

\subsubsection{Overall Performance}
Models augmented with SafePlan consistently demonstrated superior safety discrimination capabilities as seen in Fig. \ref{fig:AcceptanceRate}. Gemini 1.5 Pro with SafePlan achieved the highest accuracy (88.4\%) while maintaining a crucial balance between accepting safe tasks (70.9\%) and rejecting unsafe ones (92.9\% rejection rate). In contrast, models without SafePlan showed near-perfect recall by accepting all safe tasks but exhibited poor discrimination by accepting 61.3-85.8\% of unsafe tasks. The ethical-to-unethical acceptance ratio (E/U Ratio) most clearly demonstrates SafePlan's impact. Gemini 1.5 Pro with SafePlan achieved a ratio of 10.0 (accepting safe and ethical tasks at 10x the rate of unethical ones) compared to just 1.63 without SafePlan. Similar improvements were observed with GPT4o (9.79 vs. 1.35) and Gemini Flash 2.0 (7.47 vs. 1.17).

\subsubsection{Task Domain Analysis}
As shown in Table~\ref{tab:domain_comparison}, SafePlan's performance remained robust across different task domains. For manipulation tasks, which comprised the majority of our benchmark (456 tasks), Gemini 1.5 Pro with SafePlan achieved 91.7\% accuracy with only 4.7\% unethical task acceptance. Navigation tasks showed similar performance, while assistive tasks proved more challenging with slightly higher unsafe acceptance rates (28.9\%), likely due to nuanced ethical considerations in human-care scenarios.

\subsubsection{Impact of Task Complexity}
Table~\ref{tab:complexity_comparison} demonstrates SafePlan's effectiveness across complexity levels. Notably, SafePlan maintained high discrimination capabilities even for complex tasks, with Gemini Flash 2.0 achieving an exceptional E/U ratio of 43.2 for complex tasks. This indicates that our formal logic approach provides particular advantages for intricate scenarios where simple rule-based approaches might fail.

\subsubsection{Statistical Significance}
McNemar's test confirmed the statistical significance of SafePlan's improvements for GPT4o ($\chi^2$=14.78, p=0.0001) and Gemini Flash 2.0 ($\chi^2$=34.83, p<0.0001). While Gemini 1.5 Pro showed equivalent improvement patterns, the specific comparison did not reach statistical significance threshold (p=1.0).

Overall, these results demonstrate that our SafePlan framework significantly enhances LLM safety capabilities, achieving a significant improvement in discrimination performance compared to baseline models. This validates our approach of leveraging formal logic and structured reasoning for robotic task safety evaluation.

\subsection{Experimental Simulation Results}

Results from our experimental simulation on AI2-THOR across 18 tasks (Table~\ref{tab:performance_comparison}) showed that the Invariant COT Reasoner of the SafePlan Framework significantly improves the average Sucess Rate (SR\%) of task allocation and execution completion compared to the baseline. While baseline GPT4o achieves only 22\% success rate, the SafePlan implementations achieve much higher rates: 84\% (Gemini Flash 2.0), 95\% (Gemini 1.5 Pro), and 95\% (GPT4o). With respect to tasks for \textbf{Scene Reasoning}, the baseline executes all tasks (100\% E\%) but crashes on all of them (100\% CR\%). SafePlan models intercept most erroneous tasks before execution (0-16\% E\%). For tasks that do execute, SafePlan models avoid crashes entirely (0\% CR\%). Also, the Fault detection (F\%) is very high (84-100\%) for SafePlan, showing effective prevention. For \textbf{Skill Reasoning}, similar pattern emerges with baseline executing all tasks with 100\% crash rate. In contrast, SafePlan significantly reduces execution of problematic tasks (16-50\% E\%). Fault detection remains strong (50-84\%) and GPT4o with SafePlan particularly shows 0\% crash rate even for executed tasks. For \textbf{Language Ambiguity} baseline model performs somewhat better here (66\% SR\%) but still executes all tasks. The SafePlan models maintain high success rates (84-100\%). Interestingly, execution rates are higher in this category (84-100\% E\%), however crash rates drop to 0\% with SafePlan.

Overall, our results show that SafePlan enhances safety in LLM-based robotic systems by effectively intercepting problematic commands before execution. Either by preventing execution of tasks likely to fail (high F\%) or by ensuring tasks that do execute complete successfully (low CR\%), thus validating he effectiveness of combining formal logic with chain-of-thought reasoning for enhancing robotic safety.







\section{Conclusion}

In this research, we introduced \textit{SafePlan}, a logical framework that leverages formal logic and chain-of-thought reasoning to enhance safety in LLM-based robotic task planning. Our experimental results showed that SafePlan significantly outperforms baseline approaches, leading to a 90.5\% reduction in harmful task acceptance while still maintaining reasonable acceptance rate of ethical tasks. Future work will explore the implications of employing this framework for dynamic and complex scenarios.


\typeout{}
\bibliography{main}
\bibliographystyle{IEEEtran}
\end{document}